\documentclass[runningheads]{llncs}

\usepackage{eccv}

\usepackage{eccvabbrv}

\usepackage{graphicx}
\usepackage{booktabs}

\usepackage{url}

\usepackage{multirow}
\usepackage{amssymb}
\usepackage{pifont}
\usepackage{array} 

\usepackage{mathrsfs}  
\usepackage{bbm}
\usepackage{dsfont}
\usepackage{amsmath}
\usepackage{mathtools}
\usepackage{caption}
\usepackage{arydshln}
\usepackage{wrapfig}
\usepackage{setspace}
\usepackage{subcaption}
\usepackage[dvipsnames]{xcolor}
\definecolor{limegreen}{HTML}{DBEAC1}
\definecolor{lavender}{HTML}{EEE2F8}
\definecolor{lightorange}{HTML}{FFF7F4}
\definecolor{lightgray}{HTML}{767070}
\definecolor{lightblue}{HTML}{DAE3F4}

\definecolor{beige}{HTML}{F3E5AB}

\usepackage[accsupp]{axessibility}  

\newcommand{\mb}{\mathbf}

\newcommand{\algo}{CLIP-M$^3$ }

\usepackage{amsmath,amsfonts,bm}

\def\eqref#1{equation~\ref{#1}}

\def\1{\bm{1}}

\DeclareMathAlphabet{\mathsfit}{\encodingdefault}{\sfdefault}{m}{sl}
\SetMathAlphabet{\mathsfit}{bold}{\encodingdefault}{\sfdefault}{bx}{n}

\usepackage[pagebackref,breaklinks,colorlinks,citecolor=eccvblue]{hyperref}

\usepackage{orcidlink}

\begin{document}

\title{A streamlined Approach to Multimodal Few-Shot Class Incremental Learning for Fine-Grained Datasets} 

\titlerunning{Abbreviated paper title}

\author{Thang Doan\thanks{correspondence: thang.doan@mail.mcgill.ca} \and
Sima Behpour \and
Xin Li \and
Wenbin He\and
Liang Gou \and
\\
Liu Ren}

\authorrunning{F.~Author et al.}

\institute{Bosch Research North America $\&$ \\ Bosch Center for Artificial Intelligence (BCAI)}

\maketitle

\begin{abstract}
Few-shot Class-Incremental Learning (FSCIL) poses the challenge of retaining prior knowledge while learning from limited new data streams, all without overfitting. The rise of Vision-Language models (VLMs) has unlocked numerous applications, leveraging their existing knowledge to fine-tune on custom data. However, training the whole model is computationally prohibitive, and VLMs while being versatile in general domains still struggle with fine-grained datasets crucial for many applications. We tackle these challenges with two proposed simple modules. The first, Session-Specific Prompts (SSP), enhances the separability of image-text embeddings across sessions. The second, Hyperbolic distance, compresses representations of image-text pairs within the same class while expanding those from different classes, leading to better representations. Experimental results demonstrate an average 10-point increase compared to baselines while requiring at least 8 times fewer trainable parameters. This improvement is further underscored on our three newly introduced fine-grained datasets. 
\end{abstract}

The pursuit of Artificial General Intelligence (AGI) entails crafting models that emulate human learning, capable of seamlessly acquiring new knowledge without forgetting prior data (known as Catastrophic Forgetting, CF~\cite{catastrophic_forgetting,cf,ramasesh2021anatomy}). This endeavor aligns with lifelong learning~\cite{chaudhry2018efficient,de_Lange,aljundi_2019}, where models must adeptly accumulate insights in real-time, meticulously balancing stability and adaptability~\cite{bennani2020generalisation,mirzadeh_2020}.  However, real-life scenarios often present a challenge of limited samples from new classes, rather than a continuous stream of data. Nevertheless, the goal remains to quickly grasp new concepts. This scenario is well encapsulated by Few-Shot Class-Incremental Learning (FSCIL).

The rise of Vision-Language Models (VLMs) such as CLIP~\cite{clip} presents a promising avenue for FSCIL, capitalizing on their existing knowledge to learn from new fresh data. Yet, this advancement also brings forth new challenges. Firstly, the large scale of these models can make fine-tuning the entire network computationally prohibitive~\cite{pmlr-v232-doan23a}. Secondly, while VLMs demonstrate versatility in general domains, transposing their knowledge to fine-grained domains remains nontrivial, posing a significant challenge for many real-world applications. For instance, camera surveillance systems rely on accurate object detection~\cite{obj_detection_camera_surveillance}, autonomous driving requires real-time detection of new road traffic signs~\cite{joseph2021towards}, and agricultural crop surveillance demands precise monitoring of plants evolutions~\cite{crop_plants_ai}. 

\begin{figure}[htbp]
\vspace{-1.5em}
    \centering
     \hspace{-1.5cm}
    \begin{subfigure}{0.7\linewidth}
        \centering
        \includegraphics[width=\linewidth]{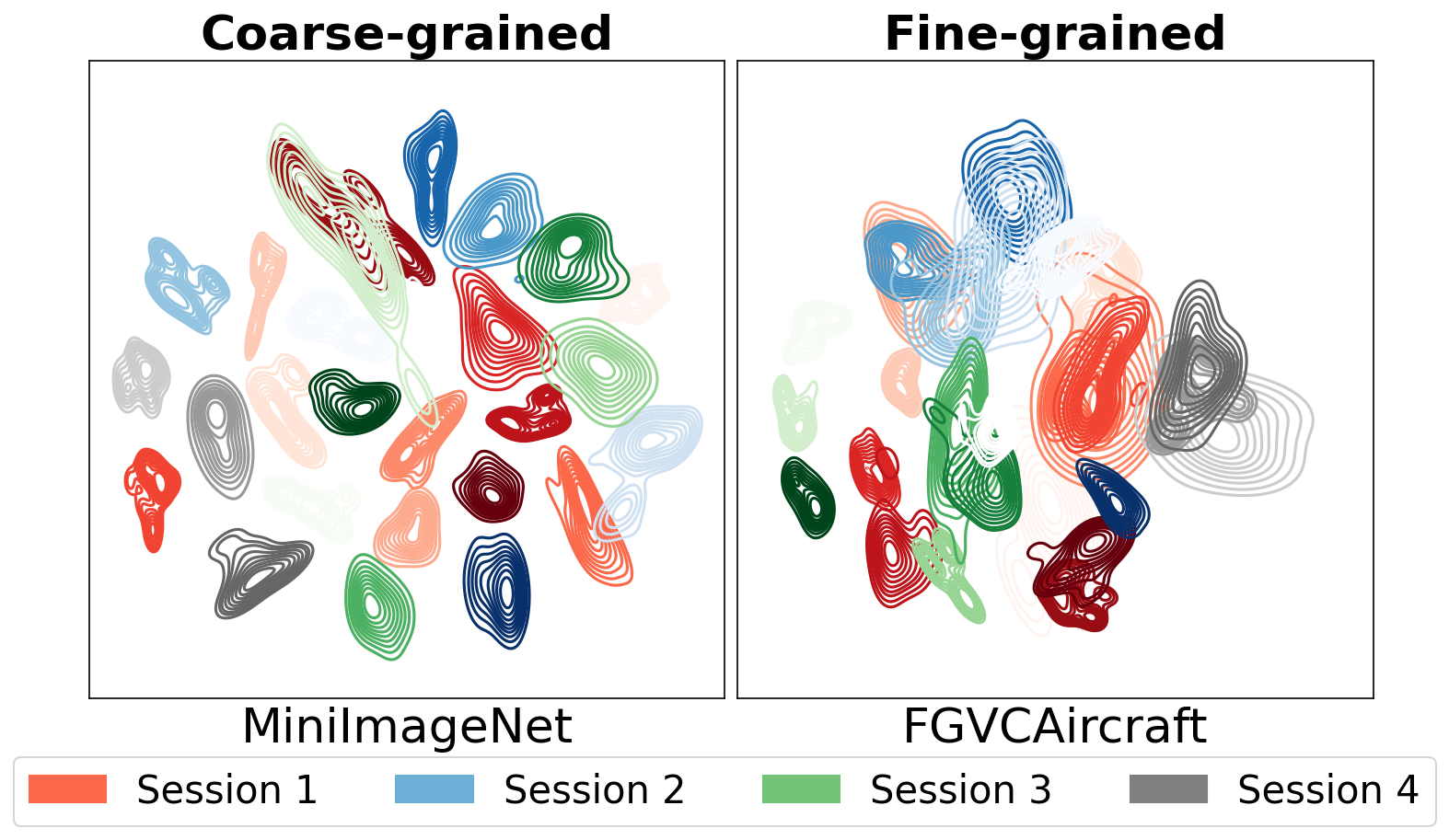} 
        \caption{\textbf{KDE plots of image embeddings, color-coded by session and class.} Coarse-grained datasets (CG, left) typically exhibit clear separability between classes, whereas fine-grained datasets (FG, right) often display closely intertwined class representations.}
        \label{fig:finegrained_vs_coarse} 
    \end{subfigure}
    \hspace{0.01\linewidth}
    \begin{subtable}{0.2\linewidth}
        \centering
        \footnotesize
         \captionsetup{justification=centering} 
         \caption{Zero-shot classification accuracy of CLIP.}
        \label{tab:zero_shot_coarse_vs_finegrained}
        \begin{tabular}{l|c}
         \toprule[1.5pt]
        Dataset & Accuracy \\
        \hline
        \text{mini}Imagenet & 82.2 \\
        FGVCAircraft & 11.8 \\
         \toprule[1.5pt]
        \end{tabular}
        \vspace{3.0cm} 
    \end{subtable}
    \vspace{-2.0em} 
\end{figure}

To address the first hurdle, recent works have proposed keeping the pretrained weights of the VLMs frozen and learning only  a limited set of parameters. This approach allows for swift adaptation to downstream tasks, either through a set of prompts~\cite{zhou2022learning,zhou2022conditional,shu2022test} or a multi-modal attention mechanism~\cite{proof,cpe_clip}.  Regarding the second challenge, despite the versatility of VLMs in general domains, they often struggle when confronted with fine-grained tasks. In such domains, where the differences between classes are narrow and highly specific, VLMs encounter difficulty in discerning intricate and subtle features essential for accurate classification. This phenomenon is evident in Table~\ref{tab:zero_shot_coarse_vs_finegrained}, where CLIP exhibits contrasting zero-shot performance on coarse (\textit{mini}ImageNet) and fine-grained (FGVCAircraft) datasets, achieving accuracies of $82.2\%$ and $11.8\%$, respectively. While CLIP possesses knowledge of airplanes, it struggles to differentiate between them. The t-SNE~\cite{t_sne} visualization (Fig ~\ref{fig:finegrained_vs_coarse}) illustrates this, where the feature representations of \textit{mini}ImageNet are distinct and highly separable, while those of FGVCAircraft (comprising 100 different airplanes) exhibit substantial overlap in feature representations.

We then propose to tackle theses challenges by positioning ourselves within the context of Parameter-Efficient Few-Shot Class Incremental Learning (FSCIL) and introducing a \underline{M}inimalist \underline{M}ulti\underline{M}odal approach, named CLIP-M$^3$. This consists of two straightforward yet potent modules:
a) the first one, Session-Specific Prompts (denoted as $SSP$), enhances the separability of text-image pairs across sessions, thereby facilitating the learning of distinct features from earlier sessions.
b) The second module leverages the curvature of the Hyperbolic distance manifold to compress the distance between image-text pairs within the same class, while exponentially expanding the distance between pairs from different classes, leading to improved representation~\cite{nickel2018learning}. As a result, our method demonstrates State-of-the-Art (SOTA) results with 10-point improvement on average while requiring at least $8$ times fewer trainable parameters. This is further validated on three newly introduced fine-grained benchmarks.

\section{Related Works}

\subsection{Few-Shot Class-Incremental Learning}
Several works have addressed Few-Shot Class Incremental Learning (FSCIL), with one stream proposing to freeze the backbone after the base session. This approach aims not only to mitigate forgetting but also to prevent overfitting on the few incoming samples~\cite{cec}. Learning of new classes is then achieved through techniques such as knowledge distillation~\cite{cheraghian2021semantic}, graph-evolving learning~\cite{cec} or neural gas topological structure~\cite{tao2020few}.
However, introducing future unknown classes can lead to confusion with existing classes and degrade performance. One solution to this challenge involves allocating or creating sufficient space~\cite{alice} in the embedding space during the base session through the creation of virtual classes~\cite{savc}, prototypes~\cite{fact}, or the generation of incremental tasks~\cite{limit}. This ensures that learning of incoming classes are sufficiently distinguishable from the base session classes.

\subsection{Vision-Language Models for Lifelong Learning}
Unlike previously mentioned Vision-Only methods that primarily operate on image embeddings, recent works harness the rich knowledge acquired from Vision-Language models such as CLIP~\cite{clip}. For instance, Ding et al.\cite{ding2022don} integrated the CLIP backbone into conventional lifelong learning methods. However, fine-tuning the entire network can be computationally expensive~\cite{pmlr-v232-doan23a} and impractical in many scenarios. To address this, recent approaches focus on training only a subset of weights or prompts~\cite{khattak2023PromptSRC, zhou2022learning}. PROOF proposes learning self-attention layers to fuse vision and text embeddings, while CPE-CLIP~\cite{cpe_clip} conditions the vision prompts on the text prompts, enabling the derivation of language-aware image features and dynamically adjusting the learning rate based on the number of samples. In contrast, IOS-CLIP~\cite{wang2023attriclip} and AttriCLIP~\cite{ios} learn class-specific attributes through a pool of key-prompt pairs. During inference, a mapping function retrieves the top-K attributes and augment the image features.

\subsection{Hyperbolic Distance}

Poincare embeddings have proven useful in learning hierarchical structures from complex symbolic or multi-relational data, such as social networks or taxonomies. This data can be represented as graphs or trees~\cite{nickel2018learning,pmlr-v97-law19a}. Thanks to its effective performance, hyperbolic distance has been applied in a wide variety of domains, including image classification~\cite{Khrulkov_2020_CVPR,yan2021unsupervised,yue2023hyperbolic,ermolov2022hyperbolic}, Natural Language Processing~\cite{sawhney-etal-2022-dmix}, and object detection~\cite{lang2022hyperbolic,ge2022hyperbolic,hypow}. In the context of image-text retrieval, Meru et al.~\cite{meru} introduced a hierarchical text structure that integrates image embeddings. However, no existing work has harnessed hyperbolic distance to address the unique challenges of Parameter Efficient Few-Shot Class-Incremental Learning with Vision-Language Model settings.

\section{Background}

\subsection{Problem Formulation}

In Few-Shot Class-Incremental Learning (FSCIL), the learner aims to acquire knowledge from a sequential data stream, denoted as $\mathcal{D}_{train}=
\{\mathcal{D}_{t}\}_{t=0}^{T}$. Each session, $\{\mathcal{D}_{t}\}=\{ (x_{i}^{t},y_{i}^{t}) \}_{i=1}^{N_t}$, includes image inputs $\bm{x}_{i} \in \mathcal{X}^{t}$ paired with labels $y_i \in \mathcal{C}^{t}$. Notably, the classes in different sessions do not overlap, which implies that $\mathcal{C}^{t} \cap \mathcal{C}^{i} = \emptyset $ for all $t \neq i$. At the end of each session $t$, the learner's performance is evaluated based on all the classes learned up to that point: $\mathcal{C}^{k \leq t }=\cup_{k=0}^{t} \mathcal{C}^{k}$. The initial session, or \textit{base session}, denoted as session $0$, is larger ($N_{0} \gg N_{t}$) compared to the subsequent incremental sessions, which contain $k$ samples per class for $n$ classes (termed \textit{n-way-k-shot}). The challenge in FSCIL lies in maintaining knowledge of the extensive base classes (dealing with class imbalance) while accommodating the limited incoming classes (addressing data scarcity)

\subsection{Efficient Prompt Learning with CLIP}

We adhere to recent literature that appends prompt tokens at the text level~\cite{zhou2022learning, zhou2022conditional} or image level~\cite{bahng2022visual}, particularly ~\cite{rasheed2023fine,khattak2023PromptSRC}, which learn both visual and text prompts separately at each transformer block, a process referred to as deep prompting.

Let us denote the image and text encoders as ${f}$ and ${g}$, respectively, and their pretrained parameters as ${\theta}_{\mathtt{CLIP}} = \{\theta_{f}, \theta_{g} \}$. Here, $\theta_{f}$ and $\theta_{g}$ represent the parameters of the image and text encoders, respectively.

The input image $\bm{x} \in \mathbb{R}^{C\times H\times W}$ is divided into $M$ patches, which are then projected to produce learnable patch tokens $\bm{e}_{cls}$. We append $V$ visual prompts, denoted as $\bm{P_{v}} = \{\bm{p_v}^1,\bm{p_v}^2, \cdots, \bm{p_v}^V\}$, resulting in the final input to the image encoder $\bm{\Tilde{I}_p}=\{ \bm{P_{v}}, \bm{e}_{cls}, \bm{e}_{1}, \bm{e}_{2}, \cdots, \bm{e}_{M}\}$. This input is used to produce the prompted visual feature $\bm{\Tilde{f}_p} = f(\bm{\Tilde{I}_p}, \theta{f})$.

Similarly, on the text feature side, we append $T$ textual prompts $\bm{P_{t}} = \{\bm{p_t}^1,\bm{p_t}^2, \cdots, \bm{p_t}^T\}$ to $\{\bm{t}_{SOS}, \bm{t}_{1}, \bm{t}_{2}, \cdots, \bm{t}_{L}, \bm{m}_{y}, \bm{t}_{EOS}\}$. Here, $\{\bm{t}_l|_{l=1}^{L}\}$ and $\bm{m}{y}$ represent the word embeddings corresponding to the text template and the class label, respectively, while $\bm{t}_{SOS}$ and $\bm{t}_{EOS}$ are the learnable start and end token embeddings. The final prompted textual features are produced as $ \bm{\Tilde{g}_p} = g(\bm{\Tilde{U}_p}, \theta{g})$, where $\bm{\Tilde{U}_p}=\{\bm{t}_{SOS}, \bm{P_{t}} ,\bm{t}_{1}, \bm{t}_{2}, \cdots, \bm{t}_{L}, m_{y}, \bm{t}_{EOS}\}$.

The visual and language prompts are collectively represented as $ \bm{P} = \{ \bm{P_{v}}, \bm{P_{t}} \}$ and are referred to as \emph{prompted features}. To simplify notation, we will omit the index corresponding to the prompt variables, leading to $\bm{\tilde{f}}$ and $\bm{\tilde{g}}$, when there is no ambiguity.

\subsection{Hyperbolic Embedding}
A Hyperbolic space is a $n$-dimensional Riemann manifold defined as $(\mathbb{B}_{c}^{n},g^{\mathbb{M}})$
with its Poincare ball $\mathbb{B}_{c}^{n}=\{ x \in \mathbb{R}^{n}: c \lVert x \rVert^{2} \leq 1 , c \geq 0   \}$ ($c$ being the constant curvature) and equipped with a Riemannian metric $(\frac{2}{1- \lVert \mb{x} \rVert^{2} })^{2}\mathbf{I}_{n}$ where $g\mathbf{I}_{n}$ is the Euclidean metric tensor. The transformation from the Euclidean to hyperbolic space is done via a bijection termed \textit{exponential} mapping $\exp_{\mb{b}}^{c}: \mathbb{R}^{n}  \rightarrow \mathbb{B}_{c}^{n}$. 
\begin{align}
\exp_{\mb{b}}^{c}(\mb{x})=\mb{b} \oplus_{c} (\tanh{(\sqrt{c}\frac{\lambda_{\mb{b}}^{c} \lVert \mb{x}  \rVert}{2})}\frac{\lVert \mb{x}  \rVert}{\sqrt{c}\lVert \mb{x}  \rVert})
\end{align}
with $\lambda_{\mb{b}}^{c}=\frac{2}{1-c \lVert \mb{b} \rVert^{2} }$ and the base point $\mb{b}$. The latter is often empirically taken as $\mb{b}=\mb{0}$ to simplify the formulas without impacting much the results \cite{ermolov2022hyperbolic}. We will also adopt this value in our study.

Inside this hyperbolic space, the distance between two points $\mb{x},\mb{y}  \in \mathbb{B}_{c}^{n}$ is computed as:
\begin{align}
d_{hyp}(\mb{x},\mb{y})=\frac{2}{\sqrt{c}}\arctan{(\sqrt{c}\lVert -\mb{x} \oplus_{c} \mb{y}  \rVert)}
\end{align}
where the addition operation $\oplus_{c} $ is defined as : \\
$\mb{x} \oplus_{c} \mb{y}=\frac{(1+2c \left< \mb{x}, \mb{y} \right> + c \lVert \mb{y}  \rVert^{2} )\mb{x} + (1 - c \lVert \mb{x}  \rVert^{2} )\mb{y} }{1+2c \left< \mb{x}, \mb{y} \right> + c^{2} \lVert \mb{x}  \rVert^{2} \lVert \mb{y}  \rVert^{2} }$.

Let's denote the projection of the embedding $\bm{\Tilde{f}}$ and $\bm{\Tilde{g}}$ into the hyperbolic embedding space as $\mb{z}=\exp_{\mb{b}}^{c}(\bm{\Tilde{f}})$ and $\mb{h}=\exp_{\mb{b}}^{c}(\bm{\Tilde{g}})$. 
When $c \to 0$, we recover the Euclidean distance: $\lim_{c \to 0} d_{hyp}(\mb{x},\mb{y}) = 2 \rVert \mb{x} - \mb{y} \rVert$. This quantity is also related to the cosine similarity $d_{cos}(\mb{x},\mb{y})=2-2\frac{ < \mb{x},\mb{y} >}{\lVert \mb{x}  \rVert \cdot \lVert \mb{y}  \rVert}$ in the case of normalized vectors (See supplementary).

For simplicity, from this point forward, we will use the variable $\bm{z}$ (respectively $\bm{h}$) to represent $\bm{\Tilde{f}}$ (respectively $\bm{\Tilde{g}}$) irrespective  of whether it is projected into the hyperbolic space.

\section{Methodology}

\begin{figure*}[h!]  
\includegraphics[width=1.0\linewidth]{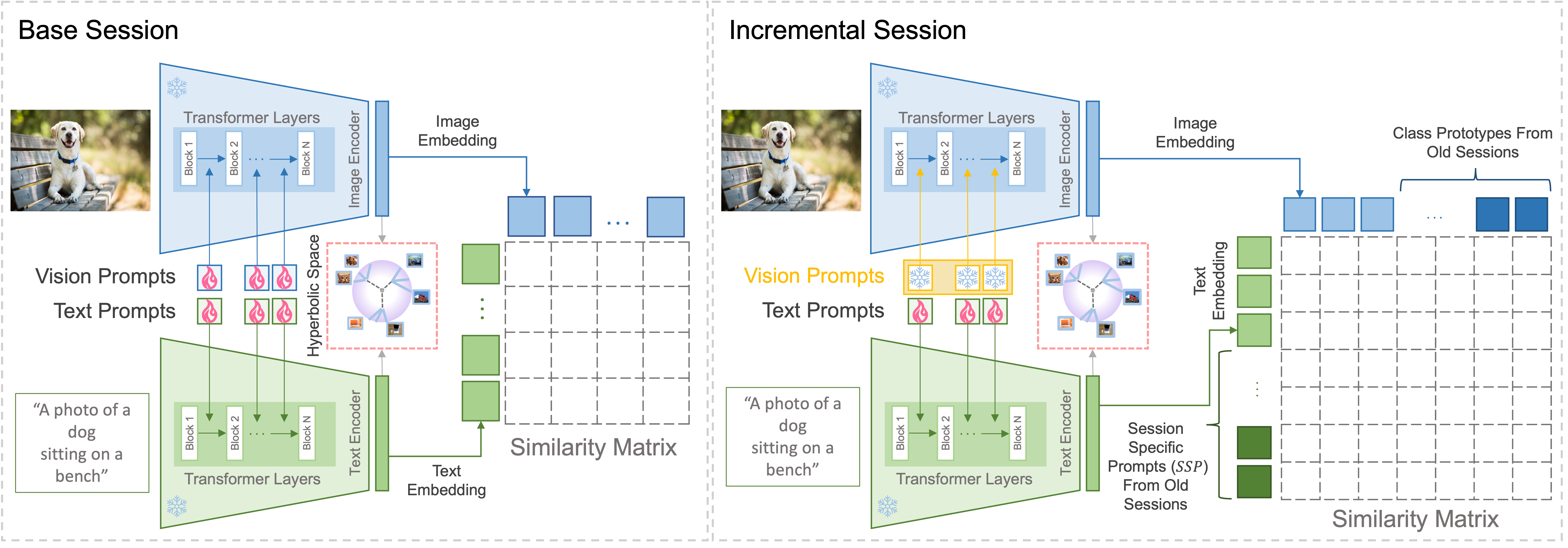}  
\scriptsize
\caption{\textbf{Overview of \algo}. During the training, the text and features prompts are interleaved within the transformer layers. Each output is then projected into Hyperbolic space and paired through a cross-entropy loss function (left). In the incremental session (right), the previously learned Session-Specific Prompts (dark green) and class prototypes (dark blues) are incorporated in the cross-entropy loss function. Note the weights of the vision prompts (yellow) being frozen and only text prompts are trained. }
\label{fig:cartoon}  
 \vspace*{-10pt}
\end{figure*}

In this section, we delve into a comprehensive description of each module of our proposed method \algo, as well as the training setting during both the base and incremental sessions. A schematic overview of these components is provided in Figure~\ref{fig:cartoon}.

\subsection{Preliminaries}

Recall that $\bm{z^{y}}$ denotes the image features and $\bm{h^{y}}$ signifies the text features, regardless of whether they are projected into the hyperbolic space or not, both associated with the label $y$. We will alternately use the terms 'image features' and 'vision prompts' where the context prevents any ambiguity.

\subsection{Base Session}
During the base session, denoted as $\mathcal{D}_{0}$, both the image and text embeddings undergo projection into the hyperbolic space. Then, they are paired using a cross-entropy loss function. We can then define the probability of the visual prompt features $\bm{z}$ belonging to class $y$ as follows:
\begin{align}
p(\bm{y}|\bm{z^{y}})=\frac{\exp(sim(\bm{z^{y}},\bm{h^{y}})/\tau)}{\sum \limits_{\bm{\tilde{y}} \in \mathcal{C}^{0}} \exp(sim(\bm{z^{y}},\bm{h^{\tilde{y}}})/\tau)}
\end{align}
where $sim(.,.)$ represents a similarity function (either cosine or hyperbolic distance) and $\tau$ a temperature.

The cross-entropy loss function for the base session is then defined as:
\begin{align}
\label{eq:LCE_base}
\mathcal{L}_{\text{CE}}^{t=0}=  \mathbbm{E}_{(\bm{z},y)\sim \mathcal{D}_{0}} [-\log p(y|\bm{z})] 
\end{align}

Following \cite{khattak2023PromptSRC}, we use a regularization strategy. This involves imposing a constraint with respect to the frozen features ($\bm{\tilde{f}}$ and $\bm{\tilde{g}}$) to prevent the model from deviating significantly from the pretrained knowledge:
\begin{align}
\label{eq:scl-features}
    \mathcal{L_{\text{image-reg}}} = \sum_{i=1}^{d}|\bm{\Tilde{f}_{p}} - \bm{\Tilde{f}}|, \; 
    \mathcal{L_{\text{text-reg}}} = \sum_{i=1}^{d} |\bm{\Tilde{g}_{p}} - \bm{\Tilde{g}}|.
\end{align}

We use textual diversity for the text prompt by using an ensemble of $M$ text templates $\bm{\tilde{g}}^{i}$ such that $\bm{\tilde{g}}=\frac{1}{M}\sum_{i=1}^{M}\bm{\tilde{g}^{i}}$. More details is discussed in the Supplementary.
The final loss to be optimized is:
\begin{align}
\label{eq:final_base}
    \mathcal{L_{\text{base}}} &= \mathcal{L}_{\text{CE}}^{t=0} + 
    \alpha \mathcal{L_{\text{image-reg}}} + \beta \mathcal{L_{\text{text-reg}}} 
\end{align}
where $\alpha$, $\beta$ are trade-off coefficients between the downstream task objective function and the regularization\footnote{Note that only the visual and text prompts parameters $\bm{P}$ are trained}.


\subsection{Incremental Sessions}

\subsubsection{Completing Base session}
 Once the base session learning completed, the user encounters incremental sessions, each containing $k$ samples from $n$ classes ($n << |\mathcal{C}^{0}|$). The objective here is to learn them without risking the loss of base session's knowledge. To achieve this, we freeze the visual prompts weights $\bm{P_{v}}$, leaving only the text features parameters $\bm{P_{t}}$ to be trainable (Figure~\ref{fig:cartoon} right, in yellow).

\subsubsection{Learning Session Specific Prompts ($SSP$)}

At the end of session $t-1$, we acquire session-specific knowledge, termed Session-Specific Prompts. Our goal is not only to preserve this knowledge but also to use it as a reference point to ensure that upcoming prompts can effectively learn features of new classes while remaining distinguishable from previous ones. To accomplish this, we save a non-differentiable copy of these prompts and incorporate them into the probability function (\textcolor{Mahogany}{brown terms}). Denoting these prompts as $\bm{\underline{h}^{y}}$, where $y \in \mathcal{C}^{k<t}$, the probability function for incremental sessions becomes:

\begin{equation}
p(\bm{y}|\bm{z^{y}}) =  \\
 \frac{\exp(sim(\bm{z^{y}},\bm{h^{y}})/\tau)}{\textcolor{Mahogany}{\sum \limits_{ \tilde{y} \in \mathcal{C}^{k<t}} \exp(sim(\bm{z^{y}}, 
  \bm{\underline{h}^{\tilde{y}}}  )/\tau)}+\sum \limits_{\tilde{y} \in \mathcal{C}^{t}} \exp(sim(\bm{z^{y}},\bm{h^{\tilde{y}}})/\tau)}, y \in \mathcal{C}^{t}
\end{equation}

Because image and text function as pairs, it is essential to consistently incorporate image features as well. To this end, after each session we save the class image prototype computed as the class average embedding ( denoted as $\bm{\underline{z}^{y}}, y \in \mathcal{C}^{k}, k<t$). Those can be stored as a separate copy (in a buffer $\mathcal{B}$) and subsequently incorporated into a cross-entropy loss function. This function is designed to preserve knowledge acquired from previous sessions by allowing the current prompts to be distinguishable from:

\begin{align}
\label{eq:LCE_former}
\mathcal{L}_{CE}^{k<t}= \mathbbm{E}_{(\textcolor{Mahogany}{\underline{z}},y)\sim \mathcal{B}^{k < t}}[-\log p(y|\textcolor{Mahogany}{\bm{\underline{z}}}) ]
\end{align}

where $\mathcal{B}^{k<t}$ represents the buffer containing all the class prototypes for sessions $k=0 \; ... \; t-1 $.

Similarly the cross-entropy loss for the current session is defined as:
\begin{align}
\label{eq:LCE_inc}
\mathcal{L}_{CE}^{t}=  \mathbbm{E}_{(z,y)\sim \mathcal{D}^{t}}[-\log p(y|\bm{z}) ]
\end{align}

While Eq~\ref{eq:LCE_former} is designed to ensure the retention of knowledge from past sessions, Eq~\ref{eq:LCE_inc} guarantees the learning of the current session by distinguishing it from previously acquired knowledge.

The final incremental session's loss can be summarized as:

\begin{align}
\label{eq:final_inc}
    \mathcal{L_{\text{inc.}}} &= \mathcal{L}_{CE}^{t} +  \gamma \mathcal{L}_{CE}^{k<t} +
    \alpha \mathcal{L_{\text{SCL-image}}} + \beta \mathcal{L_{\text{SCL-text}}}  
\end{align}
with $\gamma$ balancing the trade-off with the former sessions knowledge.

\section{Experiments}

We begin by providing an overview of our experimental setup and introduce our three newly designed fine-grained datasets. Next, we showcase a comparative evaluation of our proposed method against existing literature baselines on these established benchmarks. Finally a deep-dive analysis into each component of \algo is presented.

\subsection{Empirical Setup} 

\begin{wraptable}{r}{0.6\textwidth}
\vspace{-2.5em}
\scriptsize
 \caption{\textbf{Benchmarks statistics:}  $|\mathcal{C}^{0}|$ and $|\mathcal{C}^{t>0}|$  respectively represent the total number of classes for the base and incremental sessions. $T-1$ is the number of incremental sessions.}
  \vspace{-1em}
 \label{table:dataset}
    \begin{center}
        \setlength{\tabcolsep}{3pt}
        \renewcommand{\arraystretch}{1.1}
        \begin{tabular}{l|c|c|c|c|c}
            \toprule[1.5pt]
          Dataset &     $|\mathcal{C}^{0}|$  &  $|\mathcal{C}^{t>0}|$  & $T-1$ & k-shots & Type \\ \hline
  CIFAR100  & 60 & 40 & 8 & 5 & \textcolor{YellowOrange}{coarse} \\
  \textit{mini}ImageNet  & 60 & 40 & 8 & 5 & \textcolor{YellowOrange}{coarse}\\ 
  CUB200  & 100 & 100 & 10 & 5  & \textcolor{Green}{fine} \\
  StanfordCars  & 96 & 100 & 10 & 5 & \textcolor{Green}{fine} \\
  FGCVAircraft  & 50 & 50 & 10 & 5  & \textcolor{Green}{fine}\\
  iNF200  & 100 & 100 & 10 & 5  & \textcolor{Green}{fine}\\
            \bottomrule[1.5pt]
        \end{tabular}
    \end{center}
    \vspace{-2.5em}
\end{wraptable}
To ensure a fair comparison of models, we utilize three benchmark datasets commonly featured in the literature: CIFAR100~\cite{cifar100}, CUB200~\cite{CUB200}, and miniImageNet~\cite{miniimagenet}. We adopt the same splits as those used in previous studies such as ~\cite{tao2020few,cec,alice,savc}. To further validate our method, we introduce three additional fine-grained datasets\footnote{These datasets can be found here: \href{https://github.com/tldoan/CLIP_M3}{https://github.com/tldoan/CLIP-M3}}: StanfordCars~\cite{stanfordcars}, FGVCAaircraft~\cite{aircraft} and iNaturalist (Fungi)~\cite{inaturalist_2021} (which we refer to as iNF200 and comprises the first subset of 200 Fungi classes). Table~\ref{table:dataset} summaries statistics of all benchmarks, more detailed information can be found in supplementary section~\ref{app:dataset}.

\subsubsection{Implementation details}
The ViT-B/16 is used as a backbone for the CLIP model~\cite{clip} in our experiments and reported results averaged over 5 runs. We used deep prompting with $V=T=4$ Vision-Langauge prompts learned over the first 9 transformer layers. For the optimizer, we used SGD optimizer with momentum, by setting learning rate to $0.0025$, and a cosine annealing with warmup, for all the benchmarks. Prompts are randomly initialized with a normal distribution except the text prompts of the first layer which are initialized with the word embedding of \textit{“a photo of a \{ \}”}. All parameters such as number of epochs, learning rate, as well as details on hyperparameters tuning, can be found in the supplementary Table ~\ref{table:parameters}.

\subsubsection{Baselines} 
Our method is evaluated against existing baselines in the field, which similarly leverage Vision-Language (VL) models and train a subset of weights or prompts. we compare our approach with CPE-CLIP~\cite{ios} using identical splits. For completeness, we also include results from non-VLMs methods such as PC~\cite{xu2023flexible}, FACT~\cite{zhou2022forward} and LIMIT~\cite{zhou2022few} solely for \underline{informational purposes} rather than as direct baselines.

\subsubsection{Metrics}

In line with previous research~\cite{cec,alice,cpe_clip,cheraghian2021semantic}, we calculate and present both the average accuracy of each session $t$, denoted as $A_t$, and the overall average accuracy, represented as Avg$=\frac{1}{T}\sum_{i=1}^{T}A_t$. We also measure the performance drop, PD$=A_{0}-A_{T}$, which indicates the difference between the initial and final accuracy. This metric provides valuable insights into the model's capacity to retain information.

\begin{table*}
\scriptsize
\vspace{-1em}
 \caption{Average accuracy on all the benchmarks. PD represents different between the first and last sessions. Avg. represents the average across all the sessions. $\dagger$ denotes our reproduction.}
    \label{table:all_results}
    \vspace{-2em}
    \begin{center}
        \setlength{\tabcolsep}{1pt}
        \renewcommand{\arraystretch}{1.3}
        \begin{tabular}{lccccccccccc|cc}
            \toprule[1.5pt]
              &  \multicolumn{11}{c}{Accuracy in each session (\%)} &   \\
            \cline{2-14}

           &  \multicolumn{11}{c|}{CIFAR100}    & \multirow{2}{*}{Avg. $\uparrow$} & \multirow{2}{*}{PD $\downarrow$}   \\ \cline{2-12} 
                & 0 & 1 & 2 & 3 & 4 & 5 & 6 & 7 & 8 &  &   & &  \\   \hline

            PC~\cite{xu2023flexible}   & 76.30 & 71.89 & 67.70 & 63.40 & 60.21 & 57.31 & 55.01 & 52.79 & 50.74 &   &   &  61.71 & 25.56 \\
            LIMIT~\cite{zhou2022forward}  & 73.81 & 72.09 & 67.87 & 63.89 & 60.70 & 57.77 & 55.67 & 53.52 & 51.23 &  &  & 61.84 & 22.58  \\
            FACT~\cite{zhou2022few}  & 74.60 & 72.09 & 67.56 & 63.52 & 61.38 & 58.36 & 56.28 & 54.24 & 52.10 &  &  & 62.24 & 22.50 \\ \hdashline

             CLIP (zero-shot) & 65.8 & 63.4 & 64.0 & 63.7 & 64.2 & 64.7 & 65.2 & 64.7 & 64.6 &  &  & 64.5 &    \\ 
            CPE-CLIP~\cite{cpe_clip} & 87.83 & 85.86 & 84.93 & 82.85 & 82.64 & 82.42 & 82.27 & 81.44 & \textbf{80.52} &  &  & \textbf{83.42} & \textbf{7.31}  \\ 
             \algo (Ours) & 88.6   &  87.0 &  85.8 &  83.6 &  83.3 &  82.9 &  82.5 &          81.8 &  \textbf{80.8} &  &   &  \textbf{84.0} & \textbf{7.9}  \\ \hline \hline

           &          \multicolumn{11}{c|}{\textit{mini}ImageNet}   & \multirow{2}{*}{Avg. $\uparrow$} & \multirow{2}{*}{PD $\downarrow$}  \\ \cline{2-12} 
                 & 0 & 1 & 2 & 3 & 4 & 5 & 6 & 7 & 8 &  & &   &   \\   \hline

             LIMIT & 72.32 & 68.47 & 64.30 & 60.78 & 57.95 & 55.07 & 52.70 & 50.72 & 49.19 & & &59.05 & 23.13  \\
            PC   & 73.20 & 68.35 & 64.06 & 60.85 & 58.00 & 54.98 & 52.82 & 51.17 & 50.16 & & &59.28 & 23.04   \\
            FACT  & 72.56 & 69.63 & 66.38 & 62.77 & 60.60 & 57.33 & 54.34 & 52.16 & 50.49 &  &  & 60.69 & 22.07  \\ \hdashline

             CLIP (zero-shot) & 81.3 & 82.0 & 82.7 & 83.1 & 83.7 & 83.9 & 82.8 & 82.6 & 82.2 &  &  & 82.7 &     \\ 
            CPE-CLIP & 90.23 & 89.56 & 87.42 & 86.80 & 86.51 & 85.08 & 83.43 & 83.38 & 82.77 &  &  &86.13 & 7.46    \\ 
            \algo (Ours) &  96.0  & 95.92 & 94.5 & 94.2 & 94.2 & 93.7 & 92.8 & 92.7 & \textbf{92.5}&  &  &  \textbf{94.1} &  \textbf{3.54}    \\ \hline \hline

 \multirow{2}{*}{Method} & \multicolumn{11}{c|}{CUB200}  & \multirow{2}{*}{Avg. $\uparrow$} & \multirow{2}{*}{PD $\downarrow$} \\ \cline{2-12} 
            & 0  & 1 & 2 & 3 & 4 & 5 & 6 & 7 & 8 & 9 &  10  & &  \\ \hline
             PC & 74.06 & 70.89 & 68.13 & 63.98 & 61.54 & 58.85 & 57.56 & 55.96 & 54.28 & 53.73 & 52.40 & 61.03 & 21.66 \\
            FACT & 75.90 & 73.23 & 70.84 & 66.13 & 65.56 & 62.15 & 61.74 & 59.83 & 58.41 & 57.89 & 56.94 & 64.42 & 18.96  \\
            LIMIT & 75.89 & 73.55 & 71.99 & 68.14 & 67.42 & 63.61 & 62.40 & 61.35 & 59.91 & 58.66 & 57.41 & 65.50 & 18.48   \\  \hdashline 
            CLIP (zero-shot) & 50.1 & 49.1 & 47.8 & 45.3 & 42.2 & 45.3 & 44.1 & 43.3 & 42.0 & 43.0 & 44.4 & 45.4  &    \\ 
            CPE-CLIP & 81.58 & 78.52 & 76.68 & 71.86 & 71.52 & 70.23 & 67.66 & 66.52 & 65.09 & 64.47 & 64.60 & 70.79 & 16.98    \\
             \algo (Ours)   & 84.5 & 81.9 & 80.7 & 78.4  & 77.8  & 77.0 & 76.1 & 76.0 & 74.8 & 75.1  &  \textbf{74.9}  & \textbf{77.9}  & \textbf{9.6} \\ \hline \hline

     &          \multicolumn{11}{c|}{StanfordCars}   & \multirow{2}{*}{Avg. $\uparrow$} & \multirow{2}{*}{PD $\downarrow$}  \\ \cline{2-12} 
                 & 0 & 1 & 2 & 3 & 4 & 5 & 6 & 7 & 8 & 9 & 10 &   &   \\   \hline
        CLIP (zero-shot) & 49.8 & 52.5 & 52.5 & 52.5 & 51.7 & 52.4 & 53.9 &  54.9 & 55.9 & 55.8 & 56.7  & 53.5 &   \\ 
            CPE-CLIP$\dagger$ & 85.1 & 84.6 & 83.3 & 81.0 & 79.6 & 78.1 & 77.8 & 77.0 & 76.5 &75.3 & 75.6 & 79.4 & 9.5 \\
            \algo (Ours) &  86.1 & 86.1 & 85.2 & 83.8 & 83.4 & 83.1 & 82.7 & 82.2 & 82.3 & 81.4 & \textbf{81.7} & \textbf{83.5} & \textbf{4.2}   \\ \hline \hline
            &          \multicolumn{11}{c|}{FGVCAircraft}   & \multirow{2}{*}{Avg. $\uparrow$} & \multirow{2}{*}{PD $\downarrow$}   \\ \cline{2-12} 
                 & 0 & 1 & 2 & 3 & 4 & 5 & 6 & 7 & 8 & 9 & 10 &   &   \\   \hline
        CLIP (zero-shot) & 11.8  & 10.9  & 10.5  & 10.6  & 9.9  & 12.3  & 12.4  &12.2  & 11.5  & 11.0  & 11.8
  & 11.4 &    \\ 
            CPE-CLIP$\dagger$ & 50.9  & 44.7 & 42.5 & 39.5 & 36.2 & 38.6 & 36.6 & 36.7 &35.0 &33.6 &33.9 & 38.9 & 17.0  \\ 
            \algo (Ours) &  56.5 & 53.9 & 53.8 & 52.9 & 51.4 & 52.9 & 50.2 & 50.2 & 47.7 & 46.9 & \textbf{47.1} & \textbf{51.2} & \textbf{9.4}  \\ \hline \hline
               &          \multicolumn{11}{c|}{iNF200}   & \multirow{2}{*}{Avg. $\uparrow$} & \multirow{2}{*}{PD $\downarrow$}   \\ \cline{2-12} 
                 & 0 & 1 & 2 & 3 & 4 & 5 & 6 & 7 & 8 & 9 & 10 &   &   \\   \hline
        CLIP (zero-shot) & 4.3 & 4.8 & 5.3 & 5.1 & 5.3 & 5.9 & 5.6 & 6.1 & 5.7 &5.5 & 5.2 & 5.3 &  \\ 
            CPE-CLIP$\dagger$ & 75.4 & 68.2 & 62.8 & 58.0 & 53.8 & 50.2 & 47.0 & 44.3 & 41.7 & 39.4 & 37.5 & 52.5 & 37.4\\ 
            \algo (Ours) & 75.9 &   72.6  &  71.8 & 69.9 & 67.7 & 66.9 & 65.3 & 62.7 & 61.0 & 58.9 &  \textbf{56.9} & \textbf{66.3}  &  \textbf{19.3}  \\
            \bottomrule[1.5pt]
        \end{tabular}
    \end{center}
    \vspace{-1em}
\end{table*}

\subsection{Benchmark Results}
\noindent\textbf{CLIP Zero-Shot: Versatile in Generalist Domains, Yet Struggling with Fine-Grained Datasets.}
To delve deeper into our understanding from the introduction, we conduct a comprehensive evaluation of CLIP zero-shot capabilities across all benchmarks (as shown in Table \ref{table:all_results}). Our analysis underscores the challenges CLIP faces in handling fine-grained datasets. While CLIP zero-shot excels in general categories such as cars, birds, and airplanes—demonstrated by its performance on CIFAR100 and \textit{mini}ImageNet—it struggles to achieve satisfactory results on more specialized datasets like StanfordCars ($56.7\%$) or CUB200 ($44.4\%$). This limitation becomes particularly evident in datasets like FGVCAircraft and iNF200, where CLIP achieves accuracies lower than $12\%$. The challenge with the FGVCAircraft dataset stems from the alphanumeric labels, which provide limited information for visual feature extraction. Additionally, the low accuracy on the iNF200 dataset can be attributed to the lack of overlap with ImageNet (strict separability), a component of CLIP's pretrained data, as noted in~\cite{danish_fungi}. In summary, while CLIP zero-shot demonstrate extensive general knowledge, the method of effectively leveraging this knowledge for fine-grained domains remains an open question—one that our methods may offer a preliminary solution to.

\noindent\textbf{CLIP-M$^3$: Achieving High Performance with Minimal Parameters.}
Comparing our method with CPE-CLIP, we observe a significant improvement across all datasets, with the exception of CIFAR100 where results are comparable. \algo showcases notable enhancements ranging from a 6-point improvement on StanfordCars to approximately 20 points on iNF200 giving an average of 10-point improvement on fine-grained datasetS. It is also noteworthy that \algo achieves superior accuracy during the base session for CUB200, \textit{mini}Imagenet, FGVCAircraft and StanfordCars, accompanied by a substantial two-fold reduction in PD. 

\begin{wraptable}{r}{0.6\textwidth}
    \vspace{-2.5em}
    \scriptsize
    \setlength{\leftmargin}{30pt}
    \centering
    \caption{Our method demonstrates greater efficiency, particularly during incremental sessions where the weights of vision prompts are frozen.}
    \label{table:learnable_parameters}
    \setlength{\tabcolsep}{2pt}
    \renewcommand{\arraystretch}{1.1}
    \begin{tabular}{lcc}
        \toprule[1.5pt]
        & \multicolumn{2}{c}{\# trainable parameters}  \\
        \cmidrule(l){2-3}
        Model & Base session & Incremental sessions  \\
        \hline
        CPE-CLIP & 400K & 400K  \\
        CLIP-M$^3$ & \textbf{46k} &  \textbf{18k}   \\
        \bottomrule[1.5pt]
    \end{tabular}
    \vspace{-2em}
\end{wraptable}

From a parameter efficiency perspective, \algo demonstrates significant reduction in the number of trainable parameters required. During the base session, our approach requires 8 times fewer trainable parameters compared to CPE-CLIP (Table~\ref{table:learnable_parameters}), and nearly 20 times fewer during the incremental sessions (since the vision prompts are frozen). This efficiency arises from a distinct approach: interleaving small prompts within the transformer layers, particularly within the attention layers. This contrasts with the method employed by CPE-CLIP, which merges prompts with fully connected layers before feeding them into the vision prompts. A more detailed breakdown of the trainable parameters can be found in Section~\ref{app:trainable_params}. Moreover, we only storage in total $2 |\cup_{k=0}^{T} \mathcal{C}^{k}|$  embeddings for the class prototypes and the session specific prompts ($SSP$). This encompasses one for each vision and text class embedding, demonstrating the \textit{efficient} and \textit{minimalist} nature of our approach. Next, we delve into a comprehensive analysis to dissect and understand each component of \algo in greater detail.

\subsection{Ablation Study}

\begin{table*}
\scriptsize
\vspace{-2em}
\caption{\textbf{Effect of each component of \algo across the datasets benchmark.} Our components primarily influence fine-grained datasets (highlighted in green), with a negligible impact on coarse-grained datasets. The $SSP$ component has the most significant effect on CUB200 and StanfordCars, while the Hyperbolic Distance component exhibits the greatest impact on FGVCAircraft.} 
    \label{table:ablation}
     \vspace{-2em}
    \begin{center}
        \setlength{\tabcolsep}{3pt}
        \renewcommand{\arraystretch}{1.1}
        \begin{tabular}{l|c|c||c|c|c|c|c|c}
            \toprule[1.5pt]
         &     &  &  \textcolor{Green}{FG} &  \textcolor{Green}{FG}    &  \textcolor{Green}{FG}       & \textcolor{Green}{FG}        & \textcolor{YellowOrange}{CG}  & \textcolor{YellowOrange}{CG}  \\
          Name & Hyp. & $SSP$ &  CUB200  &  StanfordCars  & Aircraft & iNF200 & CIFAR100 & \textit{mini}ImageNet \\\cline{1-9} 
  Base  &\ding{55}  &  \ding{55}    &72.5 &   79.1  & 46.1                   &     54.8  & 80.5  &   92.9  \\
    w/o $SSP$& \ding{51} &  \ding{55}  &  73.5 &    80.3  & \textbf{47.5}    &    55.8  & 80.8 &  92.6     \\
  w/o  Hyp.  & \ding{55} &  \ding{51}   & 74.3 &    81.0  & 45.4             &     56.0  & 80.5   &    92.7  \\
Ours &\ding{51} & \ding{51}     &\textbf{74.9} &  \textbf{81.7}     & 47.1   &    \textbf{56.9}  &  80.6  &  92.4         \\
            \bottomrule[1.5pt]
        \end{tabular}
    \end{center}
    \vspace{-3.0em}
\end{table*}

\begin{figure}[h!]  
\vspace{-1.5em}
\includegraphics[width=1.0\linewidth]{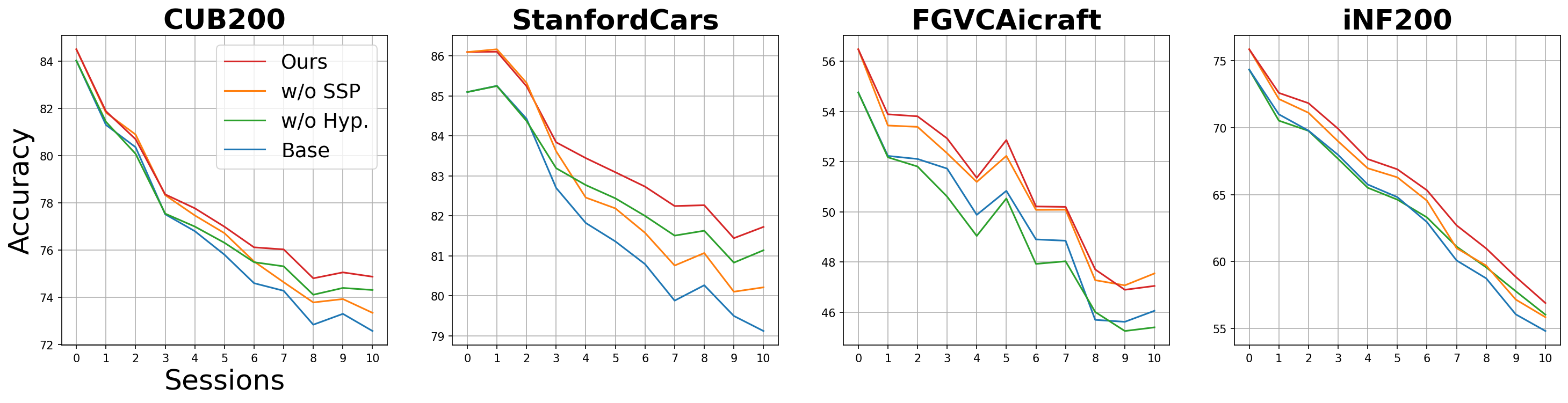}  
\caption{\textbf{Accuracy evolution across the three fine-grained datasets.}}
\label{fig:ablation_ours}  
\vspace{-2.0em}
\end{figure}

In this ablation study, we sequentially remove each component to analyze its impact on the final accuracy. The term "w/o $SSP$" refers to the strategy wherein text prompts from previous sessions keep being learned continually along with the current session's text prompt. Conversely, "w/o Hyp" refers to the use of Euclidean metrics during the calculation of the cross-entropy loss. Lastly, "Base" refers to the scenario where both components are removed. Table~\ref{table:ablation} and Figure~\ref{fig:ablation_ours} show respectively the final accuracy and its evolution across the selected benchmarks (comprehensive results are available in supplementary Table~\ref{table:ablation_all_results}). Upon examining the table, our module demonstrates significant enhancements for fine-grained datasets, aligning with the primary focus of our work. We observe improvements of $2.4\%$, $2.6\%$, $1.4\%$, and $2.1\%$ on CUB200, StanfordCars, FGVCAircraft, and iNF200, respectively. Conversely, our approach has minimal impact on coarse-grained datasets, with further discussion on this matter to follow shortly.

\begin{figure}[h!]  
 \vspace{-1.5em}
\includegraphics[width=1.0\linewidth]{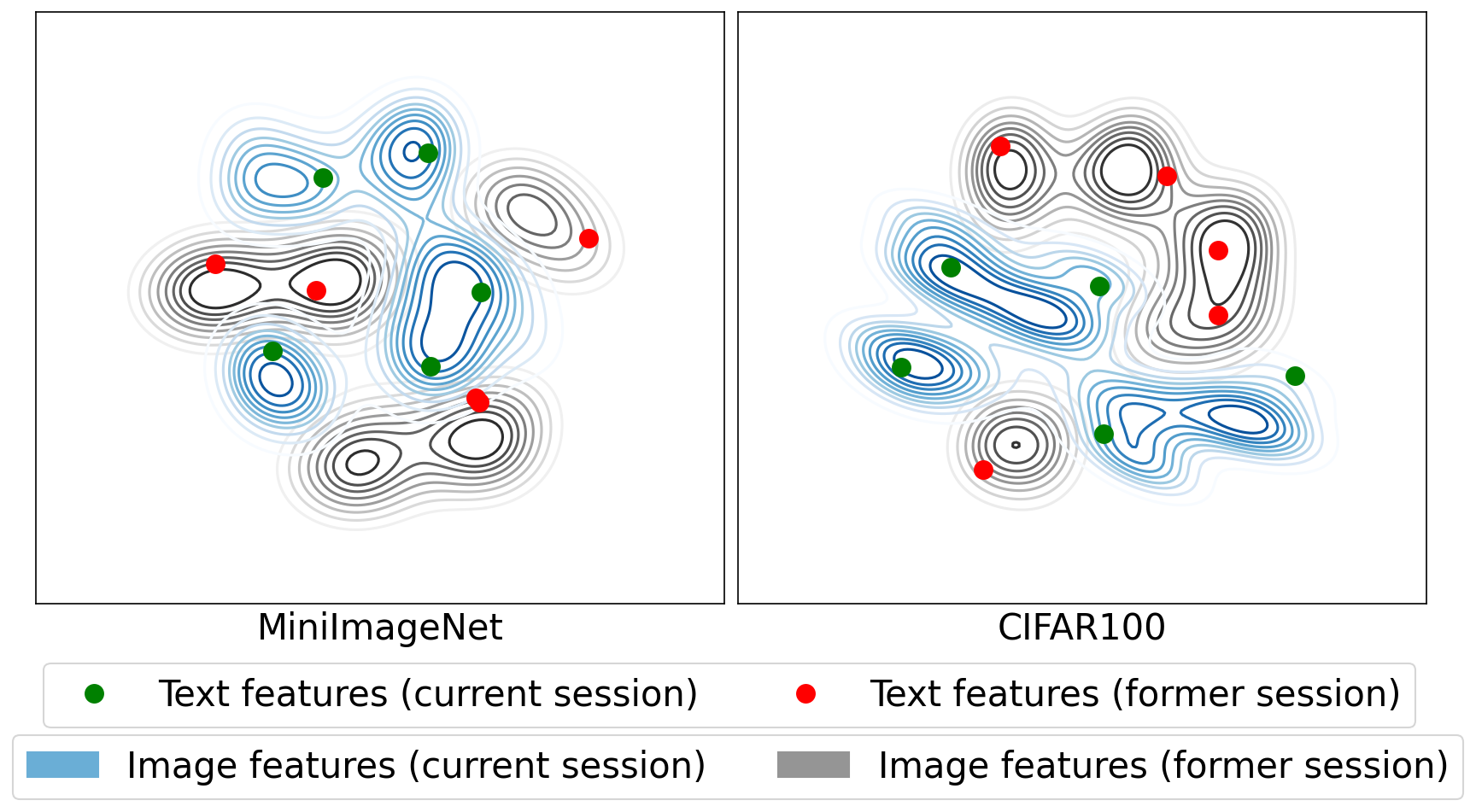}  
\caption{\textbf{Distinct Separability of Image-Text Pairs in Coarse-Grained Datasets Across Sessions.}}
\label{fig:separability_coarse_grained}  
 \vspace*{-1.5em}
\end{figure}

\noindent\textbf{Already Distinct Vision-Text Separation in Coarse-Grained Datasets.}

Investigating the marginal impact on CG datasets (CIFAR100 and \textit{mini}-Imagenet), Figure~\ref{fig:separability_coarse_grained} provides a qualitative analysis with image-text embedding (straight lineS and dot points) already notably differentiated, owing to the dataset's highly distinctive classes. This inherent separability diminishes the impact of our $SSP$ module, which is specifically crafted to augment the distinction of text prompts across sessions. Supplementary Figures~\ref{fig:s2p_impact_coarse_grained} provides additional plots supporting this argument.

\noindent\textbf{$SSP$: Enhancing Separation Between Image-Text Features Pairs Across Sessions.}

\begin{figure*}[h!]  
\vspace{-1em}
\includegraphics[width=1.0\linewidth]{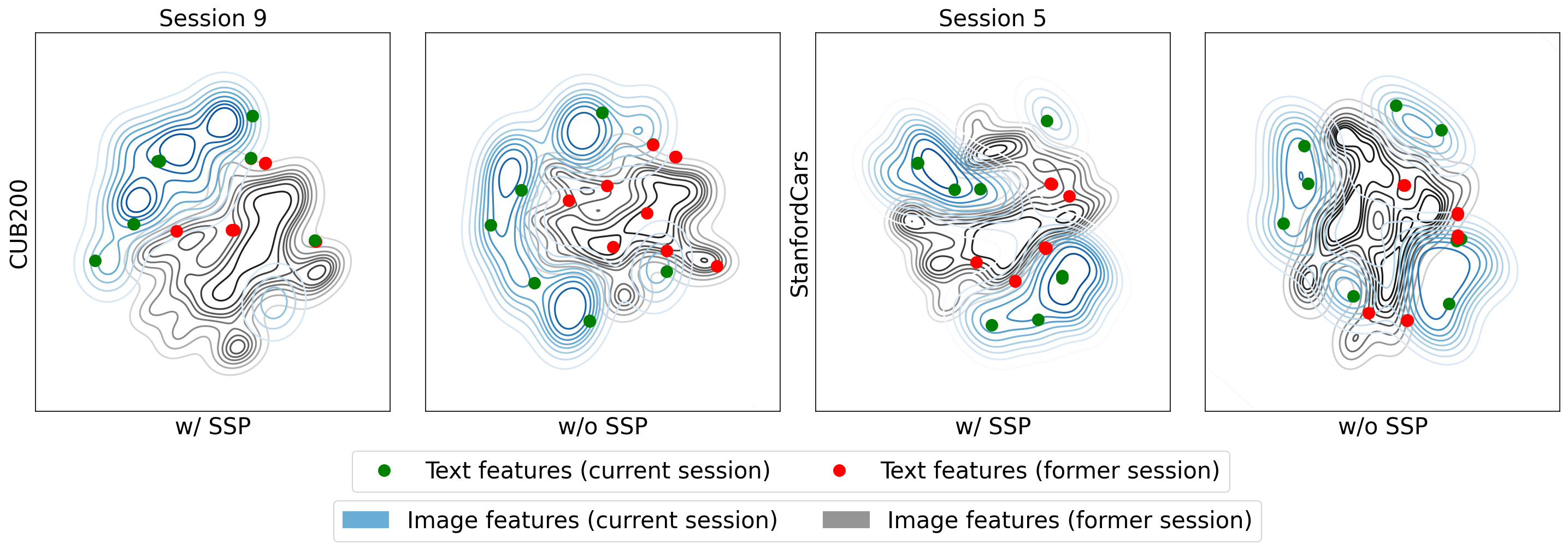}  
\caption{\textbf{Influence of $SSP$ Module on Image-Text Representation Across Sessions.} Without this module (second and fourth columns), image-text embeddings across sessions tend to be more intertwined and closely packed. Adding $SSP$ module (first and third columns) promotes a clearer differentiation and separability between sessions.}
\label{fig:FF_impact} 
     \vspace{-2em}
\end{figure*}

We now explain why $SSP$ module is effective for fine-grained datasets. As mentioned earlier in Figure~\ref{fig:finegrained_vs_coarse} (right), fine-grained datasets introduce entwined image representations, resulting from the subtle differences and minor details that differentiate one class from another. Figure~\ref{fig:FF_impact} (specifically the second and fourth columns) shows a significant overlap between the image features from different sessions (represented by blue and dark lines). This, coupled with the proximity of current session text prompts to the image feature of the previous session (indicated by red dots close to blue lines in the second column), leads to inter-session class confusion. However, the implementation of our Session-Specific Prompts (SSP) module (first and third columns) reduces that overlap between image-text pairs across sessions. It achieves this by widening the gap between the embeddings of the former and current sessions (first column for a clearer depiction of the improved separation between sessions). For clarity, Figure~\ref{fig:FF_impact} presents only the vision (represented by straight lines) and text prompt features (depicted by dots) embeddings for the current (blue) and previous (black) sessions. 
Supplementary Figures~\ref{fig:s2p_impact_cub} and ~\ref{fig:s2p_impact_stanford} offer additional plots illustrating this effect.

\noindent\textbf{The FGVCAircraft Edge Case Dataset.}

Table~\ref{table:ablation} showcases unfavorable results when applying the $SSP$ module on the FGVCaircraft dataset (third row), indicating poorer performance compared to the Base model (first row). This can be attributed to the alphanumeric nature of aircraft labels, such as '737-500' or 'A310', which constitute at least $40\%$ of the dataset. These labels lack meaningful textual information and do not correlate well with the visual features of the aircraft. While such (rare) case challenges our method, which relies on the informative value of text prompts features, it also limits the leverage of VLMs in general.

\noindent\textbf{Hyperbolic Distance: Emphasizing Image-Text Pair Proximity in Embedding Space.}

\begin{figure}[h!]  
\vspace{-1em}
\scriptsize
\includegraphics[width=1.05\linewidth]{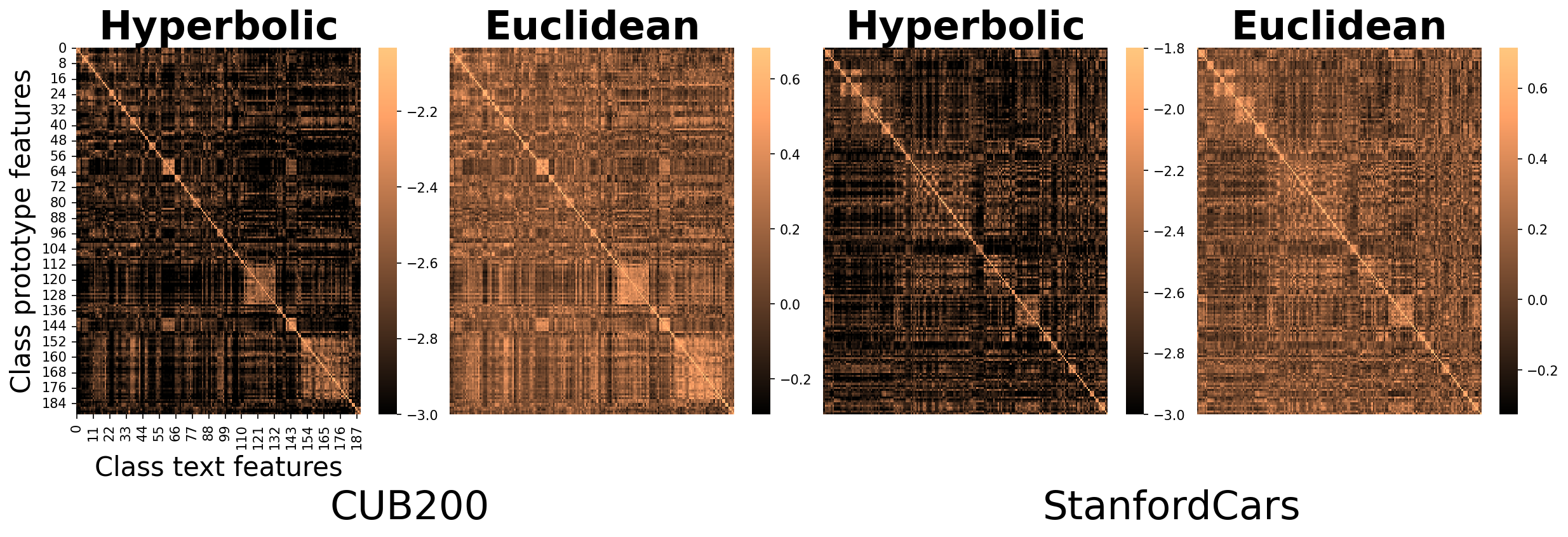}  
\caption{\textbf{Heatmap Distance between class prototype and text features for CUB200 and StanfordCars Session 9.}}
\label{fig:heatmap_hyperbolic_cosine} 
\vspace{-2em}
\end{figure}

Table~\ref{table:ablation} (second and fourth rows) showcases the impact of hyperbolic distance, contributing to at least one percent increase in accuracy over the base model. Figure~\ref{fig:heatmap_hyperbolic_cosine} provides a heatmap representation of the distance between image prototypes (y-axis) and text features (x-axis) for the CUB200 and StanfordCars dataset, with the class index on the axes. Employing Hyperbolic distance significantly intensifies the proximity between image-text pairs within the same class. This is indicated by the lighter diagonal colors, which show a stronger association between the image-text pairs of the same class compared to the use of Euclidean distance. More heatmap plots are available in Supplementary Figure~\ref{fig:heatmap_similarity_img_txt_full}.

\noindent\textbf{Impact of the Curvature Coefficient $c$.}

\begin{wraptable}{r}{0.6\textwidth}
\vspace{-2.0em}
\scriptsize
\caption{\textbf{Impact of hyperbolic curvature $c$ on final accuracy.} $c=0$ corresponds to the Euclidean distance.}
\label{table:impact_hyperbolic_curvature}
\centering
\renewcommand{\arraystretch}{1.2}
\begin{tabular}{l|cccc}
\toprule[1.5pt]
\cmidrule(l){2-3}
$c$ & 0 & 0.3 & 0.5 & 0.8   \\
\hline
CUB200 & 72.5 & \textbf{73.5} & \textbf{73.5} & 73.3  \\
StanfordCars & 79.1 & \textbf{80.3} & 80.2 & 80.2  \\
FGVCAircraft & 46.1 &   46.7 & \textbf{47.5} & 47.0  \\
iNF200 & 54.8  &  55.4  & \textbf{55.8} &  55.3 \\
\bottomrule[1.5pt]
\end{tabular}
\vspace*{-10.0pt}
\end{wraptable}
Table~\ref{table:impact_hyperbolic_curvature} quantitatively illustrates the influence of Hyperbolic curvature, denoted by $c$, which is significant across all four fine-grained datasets. An average accuracy improvement of 1-point is achieved, with remarkably stable outcomes for varying values of $c$. This stability underscores the robustness of the hyperbolic distance metric

\section{Conclusion}
In this study, we addressed the challenge of Parameter Efficient Few-Shot Class-Incremental Learning within fine-grained domains by introducing two straightforward yet impactful modules. The first, Session-Specific Prompts (SSP), enhances the separability of multi-modal features across sessions. The second module utilizes Hyperbolic distance to amplify the distance between text-image pairs within the same class while leading to a more pronounced distance between representations from different classes. This minimalist approach not only resulted in a substantial performance boost averaging 10 points on fine-grained datasets but also led to a significant reduction in trainable parameters by a factor of 8. Additionally, we introduced three novel fine-grained datasets that validated our empirical findings.

\newpage

%
%
\bibliographystyle{splncs04}
\bibliography{bibliography}

\clearpage

\section{Supplementary Material}

This supplementary material delves into the following topics:
\begin{itemize}
\item The composition of the dataset for each benchmark split
\item In-depth information regarding training and experimental procedures
\item Detailed performance results for each split
\item Count of trainable number of parameters
\item Comprehensive analysis of the impact of our $SSP$ module
\item Examination of the effect of Hyperbolic distance
\end{itemize}

\newpage

\subsection{Dataset Composition}
\label{app:dataset}

\begin{itemize}
    \item \textbf{CIFAR100} The dataset contains 60.000 images from 100 classes. We use 60 classes as the base class set. The remaining 40 classes are split into 8 sessions where each session contains 5 new classes, and the few-shot training set consists of 5 examples per class (5-way 5-shot incremental task).
    \item \textbf{\textit{mini}ImageNet} The dataset contains 60.000  images. We use 60 classes as the base class set. The remaining 40 classes are split into 8 sessions of 5 few-shot training examples each (5-way 5-shot incremental task).
    \item \textbf{Caltech-UCSD Birds-200-2011 (CUB200)} The dataset contains 11.788 images from 200 classes of bird species. We use 100 classes as the base class set. The remaining 100 classes are partitioned into 10 sessions, or timestamps, where each session contains 10 new classes, and the few-shot training set consists of 5 examples per class (10-way 5-shot incremental task).
    \item \textbf{StanfordCars} The dataset contains 16,185  images from 196 classes. We use 96 classes as the base class set. The remaining 100 classes are split into 10 sessions where each session contains 10 new classes, and the few-shot training set consists of 5 examples per class (10-way 5-shot incremental task).
     \item \textbf{FGVCAaircraft} The dataset contains 10.000 images from 100 classes. We use 50 classes as the base class set. The remaining 50 classes are split into 10 sessions where each session contains 5 new classes, and the few-shot training set consists of 5 examples per class (5-way 5-shot incremental task).
     \item \textbf{iNaturalist Fungi200 (iNF200)} This \href{https://github.com/visipedia/inat_comp/tree/master/2021}{dataset} is composed of various super categories including Plants, Insects, Birds, among others. Our focus is on the Fungi category, which is made up of $341$ distinct species. For our purposes, we use the mini images version of the dataset and select a subset of the first $200$ classes, similar to the approach used in CUB200. This subset contains 50 images per class, resulting in a total of 10,000 images. In our setup, we designate the first 100 classes as the base classes. The remaining 100 classes are then divided into 10 sessions, each containing 10 classes. Each class within these sessions includes 5 examples, resulting in a 10-way 5-shot incremental learning task.
\end{itemize} 

\clearpage

\subsection{Experimental Details}
We present below the set of hyperparameters used for each benchmarks. Note that CIFAR100 demonstrates better results without the regularzier ($\alpha=\beta=0$). We trained one GPU RTX-3090.

\begin{table*}[h!]
\scriptsize
\caption{Hyperparameters used for each benchmark.}
 \vspace{-2em}
    \label{table:parameters}
    \begin{center}
        \setlength{\tabcolsep}{3pt}
        \renewcommand{\arraystretch}{1.1}
        \begin{tabular}{l|c|c|c|c|c|c|c}
            \toprule[1.5pt]
          Parameters         &  CUB200  &  Cars  & Aircraft           &   iNF200       & CIFAR100 & \textit{mini}ImageNet & Comments \\\cline{1-8} 
   $\alpha$                    &$10$    &$10$   &  $10$              &$10$            & $0$  &$10$  &  N/A\\
     $\beta$                & $25$     & $25$   & $25$  &  $25$                       &   $0$ &  $25$ &  N/A\\
      $\gamma$                 & $30$   &   $40$ &  $20$            & $40$            & $20$ &  $30$ & $\{ 20, 30, 40\}$\\
       $c$                  &  $0.5$   &$0.8$ & $0.5$  &  $0.5$                       &  $0.5$  & $0.8$ & $\{ 0.3, 0.5, 0.8\}$ \\
        $\tau$              & $0.05$   & $0.05$  & $0.05$ &  $0.05$                   &   $0.02$ & $0.02$ & $\{ 0.01, 0.02, 0.05\}$\\
        base epochs            &  $30$  & $100$ &$100$           &$30$               &  $15$ & $5$   &  \\
        incr. epochs           & $20$  & $20$ & $10$            &  $30$              &  $10$ &  $5$ & $\{ 5, 10, 20\}$\\
        base learning rate  & $0.0025$  &$0.0025$     & $0.0025$  & $0.0025$         &   $0.0025$ & $0.0025$ & same as in ~\cite{khattak2023PromptSRC} \\
        incr. learning rate    &   $0.002$ &   $0.002$ &  $0.002$   & $0.01$         & $0.0001$ & $0.0002$ &  N/A \\
        base batch size     & $4$      &$4$    &  $32$  &  $4$                       &   $32$ & $32$ & N/A\\
        incr. batch size    &  $4$     &$4$   & $4$  & $4$                           &  $4$  & $4$ & N/A \\
        type &  FG & FG   & FG  & FG  &  CG  & CG & \\
       
            \bottomrule[1.5pt]
        \end{tabular}
    \end{center} 
\end{table*}

\subsection{Number of Trainable Parameters}
\label{app:trainable_params}
Similar to the approach in~\cite{khattak2023PromptSRC}, we integrate $V=T=4$ learnable prompts within each of the $9$ (depth size) transformer layers (See Figure~\ref{fig:cartoon}). Each vision (and text) prompt has a dimension of $[V,768]$ (and $[T,512]$, respectively), with the second dimension representing the vision width (text embeddings size respectively). Considering the $9$ transformer layers, the total number of learnable parameters amounts to $4 \times 768 \times 9 + 4 \times 512 \times 9 = 46,080$. During incremental sessions, we freeze the vision prompts and only update the text prompts, resulting in $4 \times 512 \times 9 = 28,432$ remaining learnable parameters
\clearpage

\subsection{Detailed Accuracy for each Benchmark}

\begin{table*}[h!]
\scriptsize
 \vspace{-2em}
\caption{Accuracy on the three benchmarks. PD represents different between the first and last sessions. Avg. represents the average across all the sessions.}
    \label{table:ablation_all_results}
    \begin{center}
        \setlength{\tabcolsep}{3pt}
        \renewcommand{\arraystretch}{1.1}
         \vspace{-2em}
        \begin{tabular}{lccccccccccc|cc}
            \toprule[1.5pt]
               & \multicolumn{11}{c}{Accuracy in each session (\%)} &   \\
            \cline{2-14}

        &          \multicolumn{11}{c|}{\textit{mini}ImageNet}   & \multirow{2}{*}{Avg. $\uparrow$} & \multirow{2}{*}{PD $\downarrow$}   \\ \cline{2-12} 
         & 1 & 2 & 3 & 4 & 5 & 6 & 7 & 8 & 9 &  10 & 11 & &  \\ \hline
          Base & 96.2 & 96.0 & 94.7 & 94.1 & 94.2 & 93.8 & 93.0 & 92.9 & \textbf{92.9} &   &   & 94.2 & 3.3 \\

w/o SSP & 96.0 & 95.9 & 94.5 & 93.9 & 94.0 & 93.5 & 92.8 & 92.8 & 92.6 &   &   & 94.0 & 3.4 \\
w/o Hyp & 96.2 & 96.0 & 94.6 & 94.4 & 94.3 & 93.9 & 93.0 & 93.0 & 92.7 &   &   & 94.2 & 3.4 \\
Ours & 96.0 & 95.9 & 94.5 & 94.2 & 94.2 & 93.7 & 92.8 & 92.7 & 92.5 &   &   & 94.1 & 3.5 \\\hline \hline

            &          \multicolumn{11}{c|}{CIFAR100}   & \multirow{2}{*}{Avg. $\uparrow$} & \multirow{2}{*}{PD $\downarrow$}   \\ \cline{2-12} 
               & 1 & 2 & 3 & 4 & 5 & 6 & 7 & 8 & 9 &  10 & 11 & &  \\ \hline

        Base & 88.7 & 86.9 & 85.9 & 83.7 & 83.4 & 82.8 & 82.4 & 81.7 & 80.7 &   &   & 84.0 & 8.0 \\

w/o SSP & 88.6 & 87.0 & 86.0 & 83.7 & 83.4 & 82.9 & 82.5 & 81.8 & 80.6 &   &   & 84.1 & 8.1 \\
w/o Hyp & 88.7 & 87.1 & 85.9 & 83.8 & 83.5 & 82.9 & 82.4 & 81.6 & 80.5 &   &   & 84.0 & 8.2 \\
Ours & 88.6 & 87.0 & 85.8 & 83.6 & 83.3 & 82.9 & 82.5 & 81.8 & \textbf{80.8} &   &   & 84.0 & 7.9 \\\hline \hline

             &          \multicolumn{11}{c|}{CUB200}   & \multirow{2}{*}{Avg. $\uparrow$} & \multirow{2}{*}{PD $\downarrow$}   \\ \cline{2-12} 
                 & 1 & 2 & 3 & 4 & 5 & 6 & 7 & 8 & 9 & &  & &   \\   \hline
      
            \hline
         Base & 84.0 & 81.3 & 80.4 & 77.5 & 76.8 & 75.8 & 74.6 & 74.3 & 72.8 & 73.3 & 72.6 & 76.7 & 11.5 \\
         w/o SSP & 84.5 & 81.8 & 80.9 & 78.3 & 77.5 & 76.7 & 75.5 & 74.6 & 73.8 & 73.9 & 73.3 & 77.4 & 11.1 \\
       w/o Hyp & 84.0 & 81.4 & 80.1 & 77.5 & 77.0 & 76.3 & 75.5 & 75.3 & 74.1 & 74.4 & 74.3 & 77.3 & 9.7 \\

Ours & 84.5 & 81.9 & 80.7 & 78.4 & 77.8 & 77.0 & 76.1 & 76.0 & 74.8 & 75.1 & \textbf{74.9} & 77.9 & 9.6 \\\hline \hline

 &          \multicolumn{11}{c|}{StanfordCars}   & \multirow{2}{*}{Avg. $\uparrow$} & \multirow{2}{*}{PD $\downarrow$}   \\ \cline{2-12} 
                & 1 & 2 & 3 & 4 & 5 & 6 & 7 & 8 & 9 & &  & &   \\   \hline
      
            \hline
            
Base & 85.1 & 85.3 & 84.4 & 82.7 & 81.8 & 81.4 & 80.8 & 79.9 & 80.3 & 79.5 & 79.1 & 81.8 & 6.0 \\
w/o SSP & 86.1 & 86.2 & 85.3 & 83.6 & 82.5 & 82.2 & 81.6 & 80.8 & 81.1 & 80.1 & 80.2 & 82.7 & 5.7 \\
w/o Hyp & 85.1 & 85.3 & 84.4 & 83.2 & 82.8 & 82.4 & 82.0 & 81.5 & 81.6 & 80.8 & 81.1 & 82.7 & 3.9 \\

Ours & 86.1 & 86.1 & 85.2 & 83.8 & 83.4 & 83.1 & 82.7 & 82.2 & 82.3 & 81.4 & \textbf{81.7} & 83.5 & 4.2 \\\hline \hline

 &          \multicolumn{11}{c|}{FGVCAircaft}   & \multirow{2}{*}{Avg. $\uparrow$} & \multirow{2}{*}{PD $\downarrow$}   \\ \cline{2-12} 
   & 1 & 2 & 3 & 4 & 5 & 6 & 7 & 8 & 9 & &  & &   \\   \hline
        Base & 54.8 & 52.2 & 52.1 & 51.7 & 49.9 & 50.8 & 48.9 & 48.9 & 45.7 & 45.6 & 46.1 & 49.7 & 8.6 \\
        w/o SSP & 56.5 & 53.4 & 53.4 & 52.3 & 51.2 & 52.2 & 50.1 & 50.1 & 47.3 & 47.1 & \textbf{47.5} & 51.0 & 8.9 \\
w/o Hyp & 54.8 & 52.2 & 51.8 & 50.6 & 49.0 & 50.5 & 47.9 & 48.0 & 46.0 & 45.3 & 45.4 & 49.2 & 9.4 \\

Ours & 56.5 & 53.9 & 53.8 & 52.9 & 51.4 & 52.9 & 50.2 & 50.2 & 47.7 & 46.9 & 47.1 & 51.2 & 9.4 \\ \hline \hline

   &          \multicolumn{11}{c|}{iNF200}   & \multirow{2}{*}{Avg. $\uparrow$} & \multirow{2}{*}{PD $\downarrow$}   \\ \cline{2-12} 
                 & 1 & 2 & 3 & 4 & 5 & 6 & 7 & 8 & 9 & 10 & 11  & &   \\   \hline
      
         Base & 74.3 &   71.0 & 69.8 & 68.0 & 65.8 & 64.8 & 62.9 & 60.1  & 58.7 & 56.0 & 54.8 & 64.2 & 19.5 \\
         w/o SSP & 75.9 &  72.1 & 71.1 & 69.0 & 67.0 &   66.3 & 64.6  &  61.0 & 59.7 & 57.2 & 55.8 & 65.4  & 20.0 \\
       w/o Hyp & 74.3  &   70.5  & 69.8  & 67.7    & 65.5  & 64.6 & 63.3    & 61.1  &  59.6   &57.8  & 56.0   & 64.6  &  18.3  \\

\algo (Ours) & 75.9 &   72.6  &  71.8 & 69.9 & 67.7 & 66.9 & 65.3 & 62.7 & 61.0 & 58.9 &  \textbf{56.9} & 66.3  &  19.3  \\
            \bottomrule[1.5pt]
        \end{tabular}
    \end{center}
     \vspace{-2em}
\end{table*}

In this evaluation, we examine the impact of two distinct modules: Hyperbolic Distance and Session Specific Prompts (SSP). The "w/o Hyp." scenario refers to instances where we substitute the Hyperbolic Distance with Euclidean Distance. On the other hand, the "w/o SSP" scenario refers to instances where we simultaneously learn from past text prompts in conjunction with current prompts.

Our results indicate that these modules significantly impact performance on fine-grained datasets, specifically CUB200, StanfordCars, and FGVCAircraft. In contrast, their impact on coarse-grained datasets, such as \textit{mini}ImageNet and CIFAR100, is negligible. This suggests that our modules are particularly beneficial for tasks that require a more nuanced understanding and analysis.

\clearpage

\subsection{Impact of Session-Specific Prompts ($SSP$)}
In this section, we offer a qualitative overview of our Session Specific Prompts (SSP) module. We visually demonstrate the interaction between the image features (represented by straight lines) and text features (indicated by dots) within the current (blue) and previous (red) sessions. For the sake of clarity and simplicity, we are only showcasing the interactions between the current and previous sessions. This helps us understand how these different features interact and influence each other within our SSP module.

\begin{figure*}[h!]  
\includegraphics[width=1.0\linewidth]{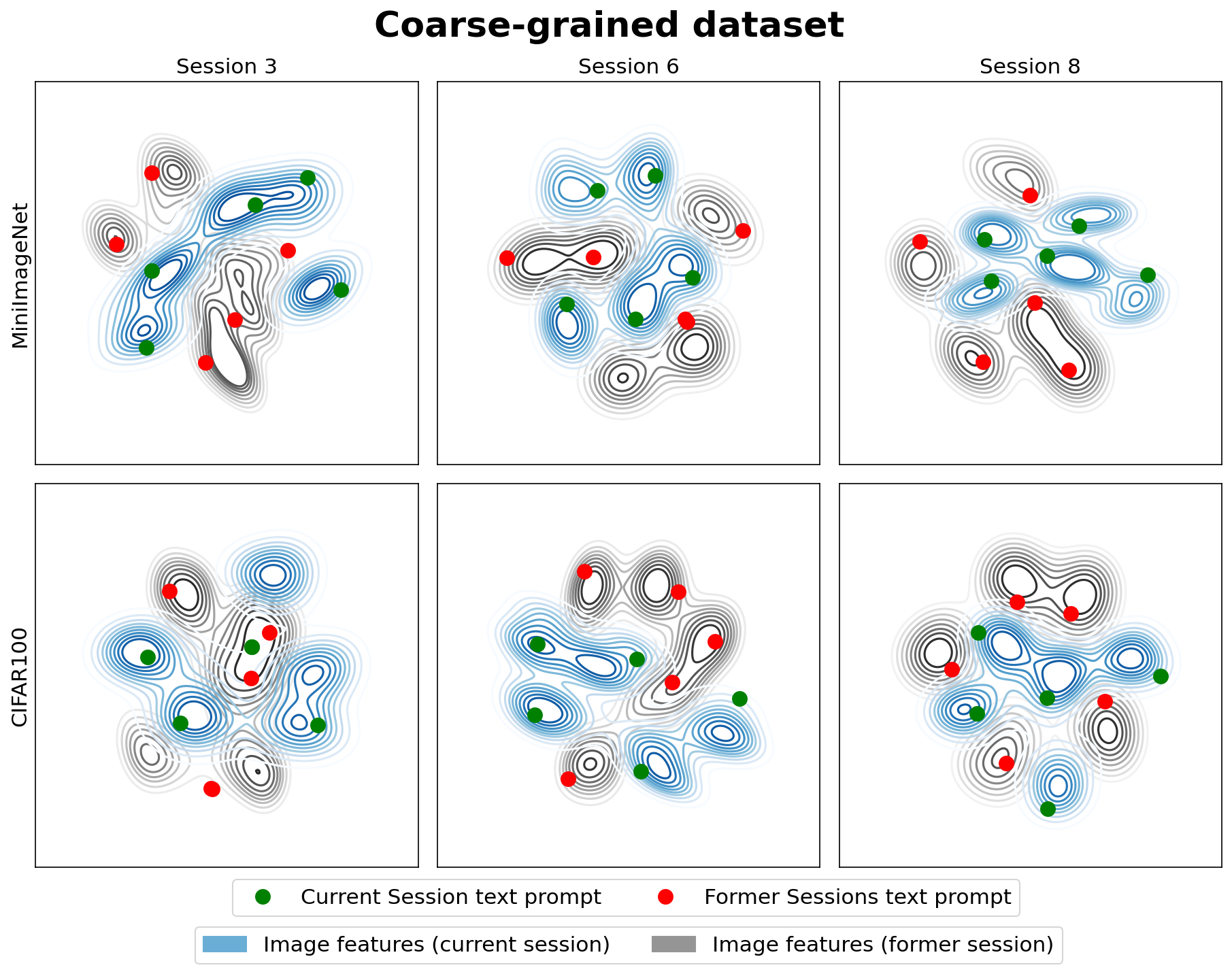}  
\caption{\textbf{Without the implementation of our SSP module, there is already a clear distinction between the image-text features of the current session (represented by blue straight lines and green dots) and those of the previous sessions (depicted by black straight lines and red dots)}}
\label{fig:s2p_impact_coarse_grained}  
\end{figure*}

Initially, we present these interactions without the influence of our module for a coarse-grained dataset (as depicted in Figure~\ref{fig:s2p_impact_coarse_grained}). As observed, the image-text features from previous and current sessions are distinctly separated. This separation results from the inherent nature of the dataset, where the classes are categorically distinct such as vehicles, animals, and furniture. This characteristic could explain why our module has a limited impact on these types of datasets. In essence, the inherent differences within coarse-grained categories may not necessitate the additional fine-tuning our module provides. Conversely, we illustrate the benefit of our module in handling high-embedding overlaps in fine-grained datasets due to intricate class details.

\begin{figure*}[h!]  
\includegraphics[width=1.0\linewidth]{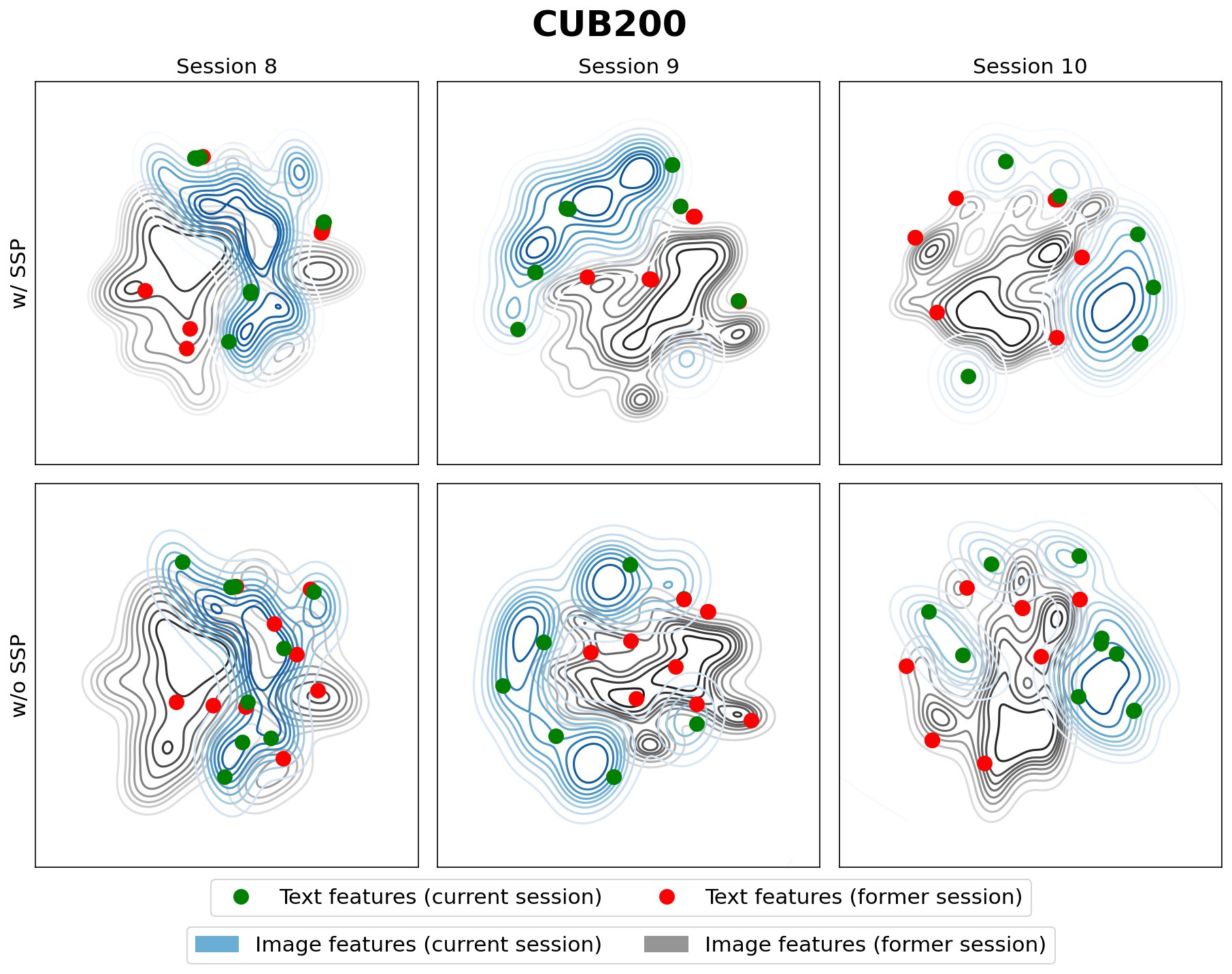}  
\caption{\textbf{Impact of the SSP module on the separation of text-image embeddings for the current session (blue lines, green dots) and the previous session (black lines, red dots).} The implementation of the $SSP$ module results in less overlap between these embedding pairs, as visible when comparing the first columns.}
\label{fig:s2p_impact_cub}  
\end{figure*}

\begin{figure*}[h!]  
\includegraphics[width=1.0\linewidth]{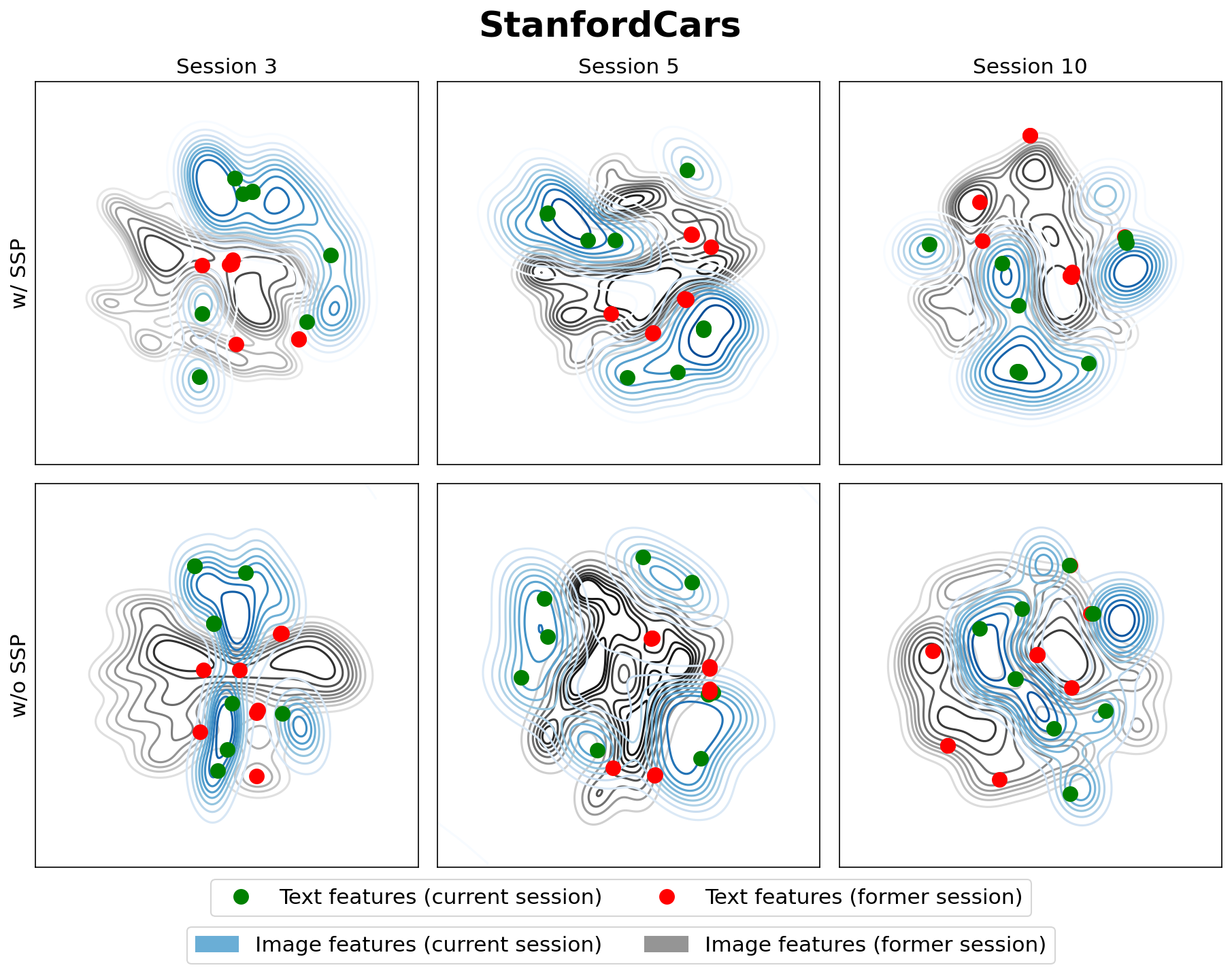}  
\caption{\textbf{Impact of the SSP module on the separation of text-image embeddings for the current session (blue lines, green dots) and the previous session (black lines, red dots).}}
\label{fig:s2p_impact_stanford}  
\end{figure*}

We now present a qualitative visualization of our $SSP$ module's impact using two fine-grained datasets: CUB200 and StanfordCars (displayed below). Without our module (as seen in the second rows), the text feature embeddings are not only densely clustered (represented by dots), but they also overlap with non-corresponding image feature embeddings (e.g., green dots in close proximity to the black straight line in Figure~\ref{fig:s2p_impact_stanford}, second row, second column). Moreover, they are also closely aligned with other text feature embeddings (green and red dots being closer can lead to class confusion as shown in Figure~\ref{fig:s2p_impact_cub}, second row, first column).

\clearpage

\subsection{Impact of Hyperbolic Distance}
Below, we assess the impact of substituting Euclidean distance with Hyperbolic distance. To achieve this, we calculate the similarity distance between the class prototype (y-axis) and text image features (x-axis) across different fine-grained datasets and different distance metric. Generally, Hyperbolic distance (first row) promotes closer proximity between image-text features within the same class (brighter color along the  diagonal), while pushing away those from different classes (darker color for off-diagonal terms), resulting in improved metric learning.
\begin{figure*}[h!]  
\includegraphics[width=1.0\linewidth]{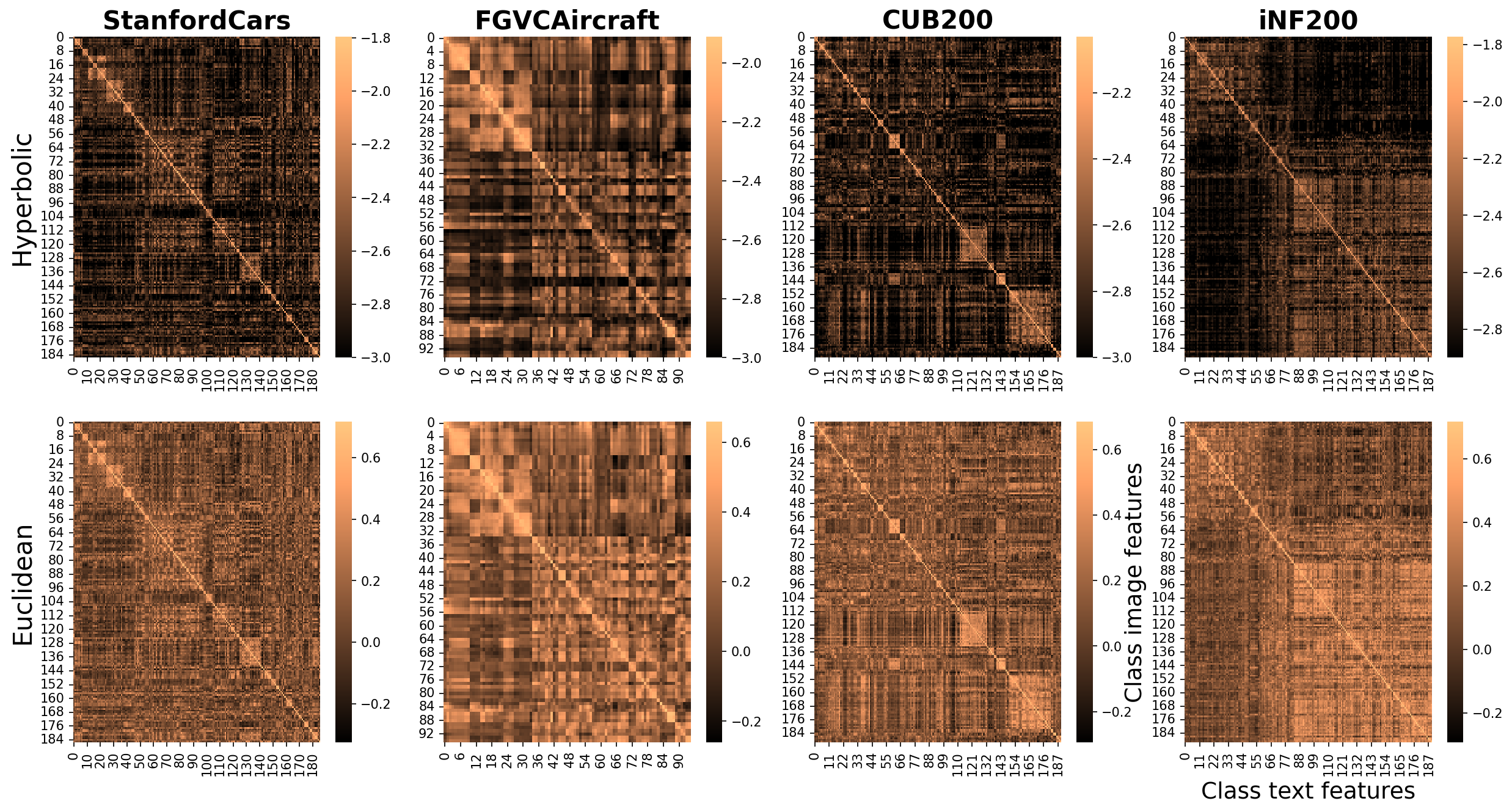}  
\caption{\textbf{Heatmap illustrating the distance between image prototypes and text features for various distances and fine-grained datasets.} Broadly, Hyperbolic Distance (first row) creates a more distinguishable separation within the same class (represented on the diagonal).}
\label{fig:heatmap_similarity_img_txt_full}  
\end{figure*} 

\end{document}